\documentclass[pdftex,preprint,12pt]{elsarticle}

\usepackage{graphicx}
\usepackage{amssymb}
\usepackage{amsthm}
\usepackage{amsmath}
\usepackage{algorithmic}
\usepackage{algorithm}

\usepackage{color}
\usepackage{comment}
\usepackage{booktabs}
\usepackage{url}

\usepackage[subrefformat=parens]{subcaption}

\usepackage{lineno}

\journal{XXX}

\begin{document}

\begin{frontmatter}

\title{Contextual Exploration Using a Linear Approximation Method Based on Satisficing}


\author[address1]{Akane Minami}
\author[address2]{Yu Kono}
\author[address2]{Tatsuji Takahashi}

\address[address1]{Graduate School of, Tokyo Denki University,  \\Ishizaka, Hatoyama-machi, Hiki-gun, Saitama 350-0394, Japan}
\address[address2]{School of Science and Engineering, Tokyo Denki University, \\Ishizaka, Hatoyama-machi, Hiki-gun, Saitama 350-0394, Japan}
\ead{tatsujit@mail.dendai.ac.jp}



\begin{abstract}

Deep reinforcement learning has enabled human-level or even super-human performance in various types of games. However, the amount of exploration required for learning is often quite large. Deep reinforcement learning also has super-human performance in that no human being would be able to achieve such amounts of exploration. To address this problem, we focus on the \textit{satisficing} policy, which is a qualitatively different approach from that of existing optimization algorithms. Thus, we propose Linear RS (LinRS), which is a type of satisficing algorithm and a linear extension of risk-sensitive satisficing (RS), for application to a wider range of tasks. 
The generalization of RS provides an algorithm to reduce the volume of exploratory actions by adopting a different approach from existing optimization algorithms. LinRS utilizes linear regression and multiclass classification to linearly approximate both the action value and proportion of action selections required in the RS calculation. The results of our experiments indicate that LinRS reduced the number of explorations and run time compared to those of existing algorithms in contextual bandit problems. These results suggest that a further generalization of satisficing algorithms may be useful for complex environments, including those that are to be handled with deep reinforcement learning.

\end{abstract}


\end{frontmatter}

\pagebreak

\tableofcontents

\pagebreak


\section{Introduction}
Reinforcement learning is a field of machine learning in which an agent learns appropriate action sequences through trial and error in the environment.
The reinforcement learning agent learns the right action sequences in a discrete state-action space using the values of state-action pairs represented in tabular form.
Deep reinforcement learning (DRL) enables the learning of appropriate action sequences in a continuous state-action space by using deep neural networks.
DRL has made remarkable progress in meeting challenges posed by complex environments \cite{DQN} \cite{AlphaGo}.
For example, robot hands that can move with human-like dexterity \cite{rl-hand} and AlphaStar that overwhelmed professional gamers in  StarCraft II, a real-time video game, \cite{alphastar} have been proposed as applications of DRL.
DRL, though, has a problem in that, when the state-action space is large, the number of exploratory actions required to learn optimal policy increases explosively.

We focused on {\em satisficing}, which is a human decision-making tendency that pursues achieving specific goals rather than acquiring the optimal action sequences.
In the complex and vast environment of the real-world, humans do not, and cannot, explore a large amount of times as that of existing optimization algorithms.
Humans are thought to be generally {\em satisficing} rather than optimizing \cite{Simon}.
Satisficing is the tendency to stop exploration, in a decision-making process, as soon as an action is found to meet a certain target level.
A previous study related to satisficing has proposed risk-sensitive satisficing (RS) \cite{RS}, which introduces satisficing into the value function in reinforcement learning.
RS can perform efficient learning with a small amount of exploration, though it depends on the target level \cite{Tamatsukuri2019}.
However, functional and linear approximation methods for RS have not been proposed; it is not easy to apply RS to deep reinforcement learning because it has no mechanism for feature-based learning.

In this paper, we propose linear RS (LinRS) as a linear approximation method of RS.
The proposed method, LinRS, linearly approximates the action value and the proportion of action selection required in the calculation of RS.
Existing optimization algorithms perform autonomous exploration using action value confidence and Bayesian formulation of the probability matching method \cite{LinUCB} \cite{LinTS}.
However, these algorithms that aim to obtain the optimal policy cannot cope with the problem of the rapid increase in the required amount of exploration as the environment becomes more complex.
Generalizations of RS enable an approach, which differs from that of typical optimization algorithms, for reducing the number of exploratory actions.
Algorithms that can find useful action sequences with a small number of explorations are especially useful in tasks such as advertisement delivery, where the number of users visiting a site is limited, and in clinical trials, where the number of patients is limited.
 
In Section \ref{sec:bandit}, we formulate multi-armed bandit problems and existing algorithms, and then introduce the RS algorithm.
In Section \ref{sec:linrs}, we propose a new algorithm, LinRS, as a generalization of RS.
In Section \ref{sec:artificial_data}, we evaluate the performance of the proposed algorithm in comparison with existing algorithms using an artificially generated dataset that can be optimized by setting a certain target level.
In the \ref{sec:real_data} section, we conduct a similar evaluation using two real-world datasets.
Based on these evaluations, in Section \ref{sec:dis}, we discuss the properties and problems when RS is extended to a linear approximation method.

\section{Multi-armed bandit problems}
\label{sec:bandit}
 
In this section, we formulate the multi-arm bandit problems and contextual bandit problems, both of which are a type of reinforcement learning tasks.
We also discuss existing algorithms for bandit problems.
\subsection{Multi-armed bandit formulation}
\label{subsec:bandit}
 
Reinforcement learning is a framework in which an agent sequentially decides actions based on past experiences and the current state.
Among the various problems in this framework, the multi-armed bandit problem is a simple task without state transitions and delayed rewards.
In the multi-armed bandit problem, a reward distribution $P=\{p_1, p_2, …, p_k\}$ is assigned to each of the $k$ choices $A =\{a_1,a_2, …, a_k\}$.
The agent obtains a reward $r_{t,a_i}$ by selecting action $a_i$ at time $t$.
The purpose of the agent in this task is to find the optimal action $a^*$ that obtains the highest reward among the $k$ choices and to maximize the cumulative reward $\sum_{t=1}^T r_{t,a_i}$ obtained by time $T$.
 
In a multi-armed bandit problem, the reward probability $p_i$ is independent of other reward probabilities.
However, this assumption does not necessarily hold for real-world tasks.
Suppose that an agent (a car company) aims to maximize the cumulative click-through rate in a car advertisement delivery task.
In this case, for the cars with the same body type, users of similar age groups and family structures tend to click on the advertisement.
This tendency can be modeled as that the reward probabilities $p_i$ are functions of the features of both the advertisement and users.
The contextual bandit problem is a typical formulation of such a setting.

The contextual bandit problems aim to maximize the cumulative reward, where the reward distributions are represented by a linear model with feature vectors.
The main difference between the types of two bandit problems is whether features to consider exist.
In the contextual bandit problem, the agent is given a $d$-dimensional feature vector ${\bf x}_{t,a}$.
The reward distribution $p_{t,a}$ is given by Equation (\ref{bandit_p})  \cite{LinUCB}.
 
\begin{equation}
\label{bandit_p}
p_{t,a} = {\bf x}^\mathrm{T}_{t,a} {\boldsymbol \theta}^*_a + \epsilon_t.
\end{equation}
${\boldsymbol\theta}^*_a$ is a vector with unknown coefficients, and $\epsilon_t$ is an error term with expectation $0$.
 
One of the most important problems in reinforcement learning is balancing exploration and exploitation.
In reinforcement learning, {\em exploration} must be performed, which involves selecting an action that has not been quite often tried yet to find the optimal action.
Nevertheless, the agent needs to carry out {\em exploitation}, that is, selecting actions that maximize the cumulative reward.
Even in the bandit problem, which is one of the simplest tasks in reinforcement learning, a balance between exploration and exploitation is difficult to achieve.
Among the two, the contextual bandit problem is particularly difficult to balance exploration and exploitation, as in that setting the reward distribution is parametrized depending on the features. Thus, the required number of exploring actions tends to be large.
 
In the bandit problem, an evaluation metric called regret is used to evaluate the balance between the exploration and exploitation of the algorithm.
Regret is defined as the difference between the cumulative action value when the optimal actions were always selected and the cumulative action value that is obtained.
It is defined by Equation (\ref{regret}) \cite{LAI_regret}.
 
\begin{equation}
\mathrm{Regret(T)} = \sum_{t=1}^T(p_{t,a^*} - p_{t,a_t}).
\label{regret}
\end{equation}
$p_{t,a^*}$ is the reward distribution for the optimal action $a^*$ at time $t$, and $p_{t,a_t}$ is the reward distribution for the selected action $a$ at time $t$.
When the optimal action is selected at time $t$, the difference is 0; the smaller regret is, the closer the agent is to the maximum cumulative reward.
The agent’s aim is to minimize regret.

\subsection{Existing optimization algorithms}
\label{subsec:existing_algo}
 
The bandit algorithm mainly aims to maximize the cumulative reward while balancing exploration and exploitation.
The most typical multi-armed bandit algorithm is the $\epsilon$-greedy algorithm.
This algorithm carries out random exploration with probability $\epsilon$ and exploitation with probability $1-\epsilon$.
The action with the highest value is called the greedy action and selected in the exploitation mode.
This algorithm has a disadvantage in that as it randomly selects actions during exploration, even apparently bad actions are selected with probability $\frac{\epsilon}{A}$.
To address this disadvantage, the upper confidence bound (UCB) algorithm \cite{UCB}, which considers the confidence of the action values and balances exploration and exploitation, is proposed.
The UCB algorithm calculates the UCB value of each action that is the sum of the action value and its UCB, and then it selects the UCB-greedy action, which is the action with the maximum UCB value.
This algorithm is based on the principle of {\em optimism in the face of uncertainty} and is characterized by the fact that it encourages the agent to aggressively explore options with low confidence.
One of the most well-known algorithms among the existing bandit algorithms is the Thompson sampling (TS) algorithm \cite{TS}.
The TS algorithm is a Bayesian formulation of the probability matching method, which matches the probability that the expected value is the maximum with the selected ratio.
It also generates posterior distributions for all alternatives and selects the action with the largest sampled value.
 
When solving the contextual bandit problem, a typical approach is to use a generalization of the multi-arm bandit algorithm to linear models.
Contextual bandit algorithms learn the estimated parameters of the action value for the feature vectors.
Typical algorithms are LinUCB \cite{LinUCB}, which is a linear extension of the UCB algorithm, and linear full posterior sampling \cite{LinTS} (hereinafter referred to as LinTS), which is a linear extension of TS.
LinUCB can efficiently calculate the value of each action when the reward distribution is linear using a closed form to reduce the computational cost of confidence intervals.
LinTS updates probability distribution as a closed form, as in a common method in Bayesian linear regression, to generate an accurate posterior distribution in a linear model.

\subsection{Satisficing algorithm: RS}
\label{subsec:rs}
 
{\em Satisficing} is a human decision-making tendency to stop exploring after finding an action that meets a certain standard \cite{Simon}.
It does not necessarily aim to find the optimal actions; instead, it seeks to find an action that exceeds a certain target level.
In this manner, algorithms based on satisficing and those based on optimization have essentially different aims.
 
RS is proposed as a value function that performs efficient satisficing when operated under the greedy policy \cite{RS}.
When the target level is exceeded by an action, the value evaluation becomes risk-aversive.
When all the actions do not exceed the target level, the evaluation becomes risk-taking.
The RS value function is defined by Equation (\ref{rs}).
 
\begin{equation}
	\mathrm{RS}_a = \frac{n_a}{N} \delta_a = \frac{n_a}{N} (E_a - \aleph).
	\label{rs}
\end{equation}
where $n_a$ is the number of times action $a$ is selected, $N$ is the total number of trials, $E_a$ is the action value obtained by action $a$, and $\aleph$ (aleph) is the aspiration level.
The decision-making algorithm that selects the greedy action that maximizes the value $\mathrm{RS}_a$ obtained by equation (\ref{rs}) is called the RS algorithm.
Hereinafter, we do not distinguish between the RS value function and RS algorithm and thus refer to both of them as RS.
In RS, the agent decides whether the action is satisfactory using the sign of $\delta_a = E_a - \aleph$ and switches its action from exploration to exploitation or vice versa.
Specifically, the agent exploits when any one of $\delta_a$ is positive and explores when all $\delta_a$ are negative.
It implies that the agent keeps exploring until it discovers an action that exceeds the aspiration level and switches to exploitation only while discovering the action.
RS is an algorithm that can dynamically resume the exploration and is robust to the update of the aspiration level and the changes in environment \cite{Hanayasu19}.
We call an action whose value exceeds the aspiration level a satisfactory action.
RS was originally derived as a generalization of the extremely symmetric form of the conditional probability (see Appendix of \cite{Tamatsukuri2019}).
Further, it has the property that the direction is the risk consideration in the value evaluation determined by the action is satisfactory.
In the case of satisfactory actions, the agent exploits in a risk-averse (pessimistic) manner by increasing the value of actions with high reliability $\frac{n_a}{N}$.
In contrast, in the case of non-satisfactory actions, the agent explores, in a risk-seeking (optimistic) manner, by increasing the value of actions with low reliability $\frac{n_a}{N}$.
 
RS can express optimization by setting the aspiration level between the largest and the second-largest reward probabilities \cite{RS_aleph}.
The optimal aspiration level is defined by Equation (\ref{aleph_opt}).
 
\begin{equation}
	\aleph_\mathrm{opt} = \frac{p_\mathrm{first} + p_\mathrm{second}}{2}.
	\label{aleph_opt}
\end{equation}
where $\aleph_\mathrm{opt}$ is the optimal aspiration level, $p_\mathrm{first}$ is the largest reward distribution, and $p_\mathrm{second}$ is the second largest reward distribution.
RS can find the optimal action sequence with a small number of exploratory actions by setting the aspiration level in accordance with equation (\ref{aleph_opt}) \cite{Tamatsukuri2019}.
This is because, when given an optimal aspiration level, the agent can only be satisfied with the action with the largest reward distribution (the optimal action).
The agent, given an optimal aspiration level, will continue to explore without being satisfied until the optimal action is discovered, and if and when it has the optimal action, it will remain satisfied and exploit the optimal action.
 
Satisficing algorithms, including RS, have two advantages in terms of their application to the real world.
The first advantage is that the business goals set by users can be easily incorporated into the algorithm.
The satisficing algorithm can find actions that can achieve the business goals while reducing the cost of the exploration by giving the business goals as aspiration levels.
The satisficing algorithm, which can change the aspiration level depending on the business goals, is more convenient than optimal algorithms, which always aim to find the optimal action.
The second advantage of satisficing algorithms is that they can demonstrate their superiority, particularly for short-term tasks.
The satisficing algorithm can reduce the number of explorations compared to optimization algorithms by setting target levels, even when it aims to find the optimal action.
For example, when humans take an exam, they would be able to study more efficiently if they are informed of the cutoff point in advance.
Similarly, a satisficing algorithm can learn efficiently by setting a target level without spending too much time on exploration.
In real-world applications, it is desirable to use an algorithm that can complete the exploration and select some useful action sequence as soon as possible, because we often have to assume that a business might be terminated or the business (and its environment) itself could change.
We propose two hypotheses based on the properties and advantages of satisficing algorithms.
By extending RS to a linear approximation, we can propose a linear approximation method that learns efficiently with a small number of explorations.
This method can also be expected to be superior in real-world tasks, where the number of explorations tends to be large, especially in early stages.
We first propose a linear approximation method for RS.
Then, using an artificial dataset with constant optimal aspiration levels, we verify whether this linear approximation method of RS has the similar properties as the ordinary RS.
In addition, using a real-world dataset, we confirm the properties of the proposed method and verify whether it is superior compared to existing optimization algorithms.

\section{Algorithm}
\label{sec:linrs}
In this section, we propose a linear approximation method for RS using the theories of linear regression and multiclass classification.
What we call Linear RS (LinRS) is an extension from the equation (\ref{rs}) to a linear function.
LinRS estimates and approximates the action value $E_a$ and reliability $\frac{n_a}{N} $ used in RS.
Because of the reason provided by LinRS, similar to RS (i.e., constant target levels as aspiration levels), aspiration levels $\aleph$ are regarded as constants without approximate estimation.
\subsection{Linear approximation of the action value}
\label{subsec:theta}
 
The agent in LinRS is given a feature vector ${\bf x}_{t,a}$ at time $t$.
Then, the agent selects an action $a_i$ and obtains a scalar reward $r_{t,a_i}$.
The linear approximation method for the action value is defined in the same manner with LinUCB \cite{LinUCB} as follows.
In the calculation of the action value, ${{\bf A}_a}$, a matrix of $d\times d$ that interaction features, and ${\bf b}_a$, a $d$-dimensional vector representing the cumulative reward that based on the features, are defined by Equations (\ref{Aa}) and (\ref{ba}), respectively.

\begin{equation}
\label{Aa}
{\bf A}_a= {\bf I}+\sum_{t=1}^{T} {\bf x}_{t,a}{\bf x}_{t,a}^\mathrm{T}.
\end{equation}
\begin{equation}
\label{ba}
{\bf b}_{a}=\sum_{t=1}^{T} {\bf x}_{t, a} r_{t, a} .
\end{equation}
The initial values of each dimension of variable vector ${\bf b}_a$ were set to zero.
The initial values of the variable matrix ${\bf A}_a$ were set to the unit matrix $\bf{I}$.
From the theory of linear regression, the update of the estimated parameter $\hat{\boldsymbol \theta}$ of the action value is defined by Equation (\ref{theta}).
 
\begin{equation}
\label{theta}
\hat{\boldsymbol \theta}_{a}={\bf A}_{a}^{-1} {\bf b}_{a}.
\end{equation}

\subsection{Linear approximation of reliability}
\label{subsec:phi}
 
LinRS uses logistic regression with the feature vector ${\bf x}_{t,a}$ as input and performs multiclass classification with each action as a class.
It also uses the probability $n_a$, which belongs to a class obtained by multiclass classification, as the percentage of trials (reliability) of each action.
The target signal $y_a$ used in the multiclass classification is generated as a weighted average of the one-hot vector $u_{t,a}$, which indicates whether each action is selected, and the baseline $\rho_{t,a}$.
The baseline $\rho_{t,a}$ is the percentage of selected actions $a$ at time $t$.
Using the baseline, we can stabilize the target signal, which is unstable when it is executed only with the variable $u_{t,a}$.
When the target signal is stable, the learning itself becomes stable, and the difficulty in estimating the approximate reliability can be mitigated.
In the linear approximation of the reliability, we use softmax as the activation function and cross-entropy as the error function.
The reliability estimate $n_a$ and the target signal $y_a$, including the baseline, are defined by Equations (\ref{n_a}) and (\ref{y_a}).
 
\begin{equation}
\label{n_a}
n_a = \mathrm{softmax}(\hat{\boldsymbol \phi}^\mathrm{T}_a {\bf x}_{t, a}).
\end{equation}
\begin{equation}
\label{y_a}
y_a = \frac{w u_{t,a} + \rho_{t,a}}{w + 1}.
\end{equation}
$\hat{\boldsymbol \phi}$ represents the estimated parameter of reliability.
Constant $w$ represents the weight of the trial at the current time, $t$.
Based on the results of the differentiation of the cross-entropy error using Equations (\ref{n_a}) and (\ref{y_a}), the update of the estimated reliability parameter $\hat{\boldsymbol \phi}$ is defined by Equation (\ref{phi}).
 
\begin{equation}
\hat{\boldsymbol \phi}_a = \hat{\boldsymbol \phi}_a + \eta (y_{a} - n_{a}){\bf x}_{t, a}.
\label{phi}
\end{equation}
The constant $\eta$ represents the learning rate.
The parameters are updated by mini-batch learning.
The data used for learning are stored in a queue.

\subsection{Linear RS}
\label{subsec:linrs} 
We define the LinRS value function, which is extended to a linear function with the same structure as equation (\ref{rs}), as expressed in Equation (\ref{f_RS}).
Given the feature vector ${\bf x}_{t,a}$, the unbiased estimator $\hat{\boldsymbol \theta}_{a}^\mathrm{T}{\bf x}_{t,a}$, which is the action value, reliability $\hat{\boldsymbol \phi}^\mathrm{T}_a {\bf x}_{t,a}$, and aspiration level $\aleph$.

\begin{equation}
\label{f_RS}
f_{a}^{\mathrm{RS}}=\mathrm{softmax}(\hat{\boldsymbol \phi}_{a}^{\mathrm{T}} {\bf x}_{t, a})\left(\hat{\boldsymbol \theta}_{a}^{\mathrm{T}} {\bf x}_{t, a}-\aleph\right).
\end{equation}
Algorithm \ref{alg1} shows the details of the entire LinRS algorithm with input parameters $\aleph, w, \eta$.
 
\begin{algorithm}[H]
	\caption{LinRS Algorithm}
	\label{alg1}
	\begin{algorithmic}[1]
	\REQUIRE $\aleph \in \mathbb{R}, w \in \mathbb{R}_+, \eta \in \mathbb{R}_+$
	\FOR{$t=1,2, ..., T$}
	\STATE Observe features of all arms $a \in A_t : {\bf x}_{t,a} \in \mathbb{R}^d$
	\STATE Initialize, ${\bf A}_a \leftarrow {\bf I}_d$  ${\bf b}_a \leftarrow 0_{d\times 1}$ $c_a \leftarrow 0$ for all $a \in A_t$
	\FOR{all $a \in A_t$}
	\STATE $\hat{\boldsymbol \theta}_a \leftarrow {\bf A}_a^{-1} {\bf b}_a$
	\STATE $f_{a}^{\mathrm{RS}}\leftarrow\mathrm{softmax}(\hat{\boldsymbol \phi}_{a}^{\mathrm{T}} {\bf x}_{t, a})\left(\hat{\boldsymbol \theta}_{a}^{\mathrm{T}} {\bf x}_{t, a}-\aleph\right)$
	\ENDFOR
	\STATE Pull arm $a_i = \mathrm{argmax}_{a \in A_t} f_{a}^{\mathrm{RS}}$
	\STATE Observe the reward $r_t$
	\STATE ${\bf A}_{a_i} \leftarrow {\bf A}_{a_i} + {\bf x}^\mathrm{T}_{t,a_i} {\bf x}_{t,a_i}$
	\STATE ${\bf b}_{a_i} \leftarrow {\bf b}_{a_i} + r_t {\bf x}^\mathrm{T}_{t,a_i}$
	\STATE $c_{a_i} \leftarrow c_{a_i} +1$
	\FOR{all $a \in A_t$}
	\STATE $u_a \leftarrow \left\{                    	\begin{array}{l}
                        	1 \: (a = a_i) \\
                        	0 \: (\mathrm{otherwise})
                    	\end{array}
                	\right.$
    	\STATE $\rho_a \leftarrow \frac{c_a}{t}$
	\STATE $n_a \leftarrow \mathrm{softmax}(\hat{\boldsymbol \phi}^\mathrm{T} {\bf x}_{t,a})$
	\STATE $y_a \leftarrow \frac{w u_{a} + \rho_a}{w+1}$
	\STATE $\hat{\boldsymbol \phi}_a \leftarrow \hat{\boldsymbol \phi}_a + \eta (y_a - n_a){\bf x}_{t,a}$
	\ENDFOR
	\ENDFOR
	\end{algorithmic}
\end{algorithm}
The decision-making algorithm, which selects the greedy action with the maximum value function $f_{a}^{\mathrm{RS}}$, is called the LinRS algorithm.
Hereinafter, we do not distinguish between the LinRS value function and the LinRS algorithm and thus refer to both of them as LinRS.
LinRS can reduce the number of explorations compared to existing optimization algorithms, provided that it retains the properties of RS.
Because LinRS can efficiently learn satisfactory actions while reducing the number of explorations, it is effective when a large number of trials cannot be performed due to environmental constraints.
For example, in news recommendation systems wherein the number of users visiting a site is limited, it is advantageous to discover useful actions with a small number of explorations.
In fact, contextual bandit algorithms are gaining popularity in news recommendation systems, such as Digg and Yahoo! Buzz \cite{Li2011}. 
Moreover, LinRS can be widely applied to services that utilize recommendation and personalization because it can consider the individuality of users, which is difficult to perform with RS. 

The optimal aspiration level of LinRS calculated from equation (\ref{aleph_opt}) is unlikely not always constant.
This problem is caused by the optimal aspiration level being defined from the reward distribution, which changes depending on the feature vector.
It is not easy for LinRS, though similar with RS, to provide the optimal aspiration level even when the reward distribution is clear.
However, it is not a severe problem.
Satisficing algorithms are useful not only when aiming to find the optimal action sequence initially.
Satisficing algorithms can show their superiority in many real-world tasks, where the cost of exploration is limited by being given the business goal as the aspiration level.
Even if the optimal aspiration level is given, LinRS does not always guarantee optimality owing to the influence of the approximation error.
In this study, therefore, the aspiration level of LinRS is given as a constant, not as a value that changes depending on the feature vector.

\subsection{Comparison with related algorithms}
\label{subsec:Comparison with related algorithms}

The TS algorithm and LinTS are used as standard baselines because they are easy to implement and show superior performance \cite{LinTS} \cite{TS_yahoo}.
However, these two algorithms have a problem because they require more runtime for decision-making than other existing algorithms.
This problem is caused by the fact that posterior distributions take time to generate.
In real-world tasks, where quick decision-making is required, this problem of runtime consumption in decision-making is often a problem.

In real-world problems with complex features and complex reward distributions, optimization algorithms, such as LinUCB and LinTS, can perform efficient exploration. However, reducing the amount of exploration is difficult compared to satisficing algorithms. 
As optimization algorithms aim to find optimal actions, the number of explorations increases with the complexity of the environment. 
However, LinRS can reduce the influence of the complexity of the environment as it aims to find satisficing actions that exceed the target level. 
Furthermore, the satisficing algorithm does not preclude optimization. Rather, by setting the optimal aspiration level, it can aim at finding optimal actions as more surely than optimization algorithms \cite{Tamatsukuri2019}. 
Whether to seek optimal actions or satisficing actions depends on the goal of the task. 
Thus, LinRS is more flexible than LinUCB and LinTS because it can change the target action according to various objectives. 

\section{Experiments with Artificial Datasets}
\label{sec:artificial_data}
 
In this section, we evaluate the performance of the proposed method, LinRS, on a contextual bandit problem using a dataset sampled from an artificial generative distribution (hereinafter referred to as the artificial dataset).
Using the contextual bandit problem, which is a simple linear approximation task, we focus on the change in the properties of LinRS owing to the linear approximation extension.
We also designed an experiment that could be linearly approximated by artificially generating the reward distribution. 
Unlike real datasets, artificial datasets are guaranteed to have a linear approximation of the reward distribution. 
Therefore, an artificial dataset is less affected by its complexity, which simplifies the analysis by focusing on the characteristics of LinRS. 
Furthermore, the guarantee of linear approximation is important because the contextual bandit problem assumes that the reward distribution is represented by a linear model using feature vectors. 

In this experiment, to compare LinRS and existing optimization algorithms using the same evaluation metrics, we used an artificial dataset in which the optimal aspiration level $\aleph_\mathrm{opt}$ of LinRS is always constant.
Given the optimal aspiration level, LinRS and existing optimization algorithms can both be defined as algorithms that aim to determine the optimal action.
By setting the same goal, the LinRS and existing optimization algorithms with the same evaluation metrics can be compared.
Moreover, using an artificial dataset in which the optimal aspiration level is always constant, LinRS can, in the same manner with RS, provide the optimal aspiration level as a constant.
To focus on the change in the properties of LinRS depending on its linear approximate expansion, providing the aspiration level to RS and LinRS in the same manner is important.
 
\subsection{Artificial dataset}
\label{subsec:artficial_data_set}
We define the reward distribution from a linear model that uses the feature vector, parameters of the reward distribution, and the error term.
The dimension of the feature vectors is $d = 128$, and the number of actions is $k = 8$.
The feature vector ${\bf x}_{t,a}$ at time $t$ is given as table data generated from uniform random numbers with dimensions $0, 1$.
The parameters of the reward distribution ${\boldsymbol \theta}^*_a$ are defined by equation (\ref{data_param}), using a diagonal matrix with mean $\mu = 0$ and covariance of $\sigma = 0.01$ that of the diagonal component only.
 
\begin{equation}
\label{data_param}
{\boldsymbol \theta}^*_a \sim \mathcal{N}(0, \sigma \times  {\bf I}).
\end{equation}
The error term $\epsilon_t$ is sampled at each time $t$ from a normal distribution $\mathcal{N}(0, 0.1)$ with mean $\mu = 0$ and variance $\sigma^2 = 0.1$.
The reward distribution for each action is defined by Equation (\ref{data_p}).
 
\begin{equation}
\label{data_p}
p_{t,a} = \mathrm{sigmoid}({\bf x}^\mathrm{T}_{t,a} {\boldsymbol \theta}^*_a + \epsilon_t).
\end{equation}
\subsubsection{Dataset with a constant optimal aspiration level}
To compare LinRS and existing optimization algorithms with the same evaluation metrics, we generated an artificial dataset in which the optimal aspiration level $\aleph_\mathrm{opt}$ of LinRS is always constant.
In this experiment, we set the optimal aspiration level $\aleph_\mathrm{opt} = \{0.5, 0.7, 0.9\}$ and generated three types of artificial datasets.
In general, it is difficult for LinRS to always set the optimal aspiration level $\aleph_\mathrm{opt}$ in contextual bandit problems where the reward distribution changes.
 
We created a dataset with a constant optimal aspiration level using the following procedure:
\begin{enumerate}
\renewcommand{\labelenumi}{(\arabic{enumi})}
  \setlength{\parskip}{0cm}
  \setlength{\itemsep}{0.1cm}
\item Generate an arbitrary parameter ${\boldsymbol \theta}^*_a$ using equation (\ref{data_param}).
\item Generate a large number of feature vectors ${\bf x}_{t,a}$ for the arbitrary parameter ${\boldsymbol \theta}^*_a$.
\item Extract only feature vectors ${\bf x}_{t,a}$ and reward distributions $p_{t,a}$ with $p_\mathrm{first} > \aleph_\mathrm{opt} > p_\mathrm{second}$.
\end{enumerate}
By repeating this procedure, we obtain $n = \{550297, 940650, 117046\}$ data for each $\aleph_\mathrm{opt}$.
\subsection{Experimental setup}
\label{subsec:artficial_data_setup}
 
We conducted experiments using these three different datasets.
We considered one step to be the process of the agent selecting an action in accordance with the algorithm and ran it for 100,000 steps.
Further, we consider one simulation to be the process of an agent completing all these 100,000 steps.
The average of regret over 100 simulations is regarded as the result.
Regret is a measure to evaluate the balancing of exploration and exploitation of the algorithm. 
A low regret indicates that the algorithm can utilize good choices while reducing the number of explorations. 
When the algorithm is stuck in a suboptimal action (local solution), regret increases linearly. 
Meanwhile, regret increases logarithmically when the optimal action is found by exploration and that action is exploited. 
In this experiment, we evaluate the balancing between exploration and exploitation of each algorithm from the transition of regret and then evaluate whether each algorithm can select good actions. 
The number of steps was set to 100,000 steps, which is sufficient to obtain stable results in the increasing tendency of regret and the relationship between the size of regret while confirming the early exploration tendency.
To compare the calculating time of the algorithms, we also measured the average execution time per simulation.
The agent selected all actions 10 times immediately after the start of the simulation to update the parameters of each algorithm.
We set $\mathrm{batch size} = 20$ for all the algorithms.
As the parameter update of the reliability of LinRS does not use a closed form, we set $\mathrm{epoch} = 5$ only for the parameter update of the reliability.
The maximum length of the queue of LinRS is set to 100.
 
\subsection{Evaluation metric: greedy rate}
\label{subsec:greedy}
In these experiments, we used the greedy ratio, which is the ratio of greedy actions selected, as an evaluation metric, in addition to regret.
The greedy ratio is a metric that lets us intuitively grasp the autonomy of the agent's exploration.
Because the agent mainly selects greedy actions during exploitation, a greedy ratio close to 1 implies that the agent has performed exploitation, and a greedy ratio close to 0 implies that the agent has performed exploration.
In the calculation of the greedy ratio, at each time $t$, a value of 0 is given if the agent selects an action other than the greedy action, and 1 if the agent selects a greedy action.
The greedy rate can be calculated by calculating the average at each time $t$ after running the entire simulation.
\subsection{Algorithm}
\label{subsec:artficial_data_algo}
 
The proposed algorithm, LinRS, was compared and verified with the existing algorithms, LinUCB, and LinTS, which are commonly used in contextual bandit problems.
We tuned the parameters of all the algorithms.
  	\begin{itemize}
  	\item \textbf{LinUCB} : This algorithm is an extension of UCB to linear functions.
It selects an action using a value function, which is the unbiased estimate of the action value plus $\alpha$ times the unbiased estimate of the variance.
In these experiments, we set the parameter $\alpha = 0.1$.
\item \textbf{LinTS} : This algorithm is an extension of TS to linear functions.
The parameter of the action value $\hat{\boldsymbol \theta}_a$ is generated from a multivariate normal distribution using vectors and matrices calculated from $\lambda$.
The variance scaling parameter $\sigma^2$ is generated from the inverse gamma distribution using $a_t, b_t$.
In this experiment, we set the parameters as $\lambda = 0.25, a_0 = b_0 = 6$.
\item \textbf{LinRS}: This is the proposed algorithm shown in Algorithm \ref{alg1}.
In this experiment, we set the aspiration level $\aleph = \{\aleph_\mathrm{opt}-1, \aleph_\mathrm{opt}-0.5, \aleph_\mathrm{opt}\}$ based on the optimal aspiration level $\aleph_\mathrm{opt} = \{0.5, 0.7, 0.9]\}$ for each artificial dataset.
Because of the possibility of underestimating the reward expectation of satisfactory actions due to the effect of the linear approximation error, we also prepared an aspiration level that was slightly lower than the optimal aspiration level.
The parameter required for learning reliability was set to $w = \eta = 0.1$.     
  	\end{itemize}
\subsection{Results}
\label{subsec:artficial_data_result}
\begin{figure}[htbp]
	\begin{tabular}{cc}
  	\begin{minipage}[t]{0.49\hsize}
    	\centering
    	\includegraphics[keepaspectratio, scale=0.2]{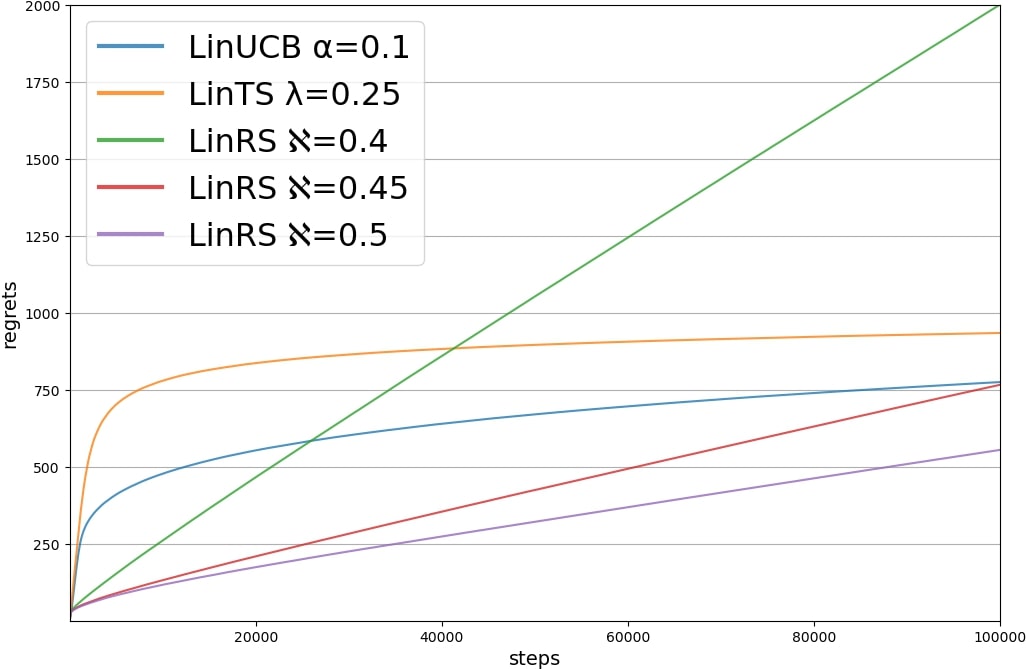}
    	\subcaption{regret: $\aleph_\mathrm{opt}$ = 0.5}
    	\label{regret_0.5}
  	\end{minipage} &
  	\begin{minipage}[t]{0.49\hsize}
    	\centering
    	\includegraphics[keepaspectratio, scale=0.2]{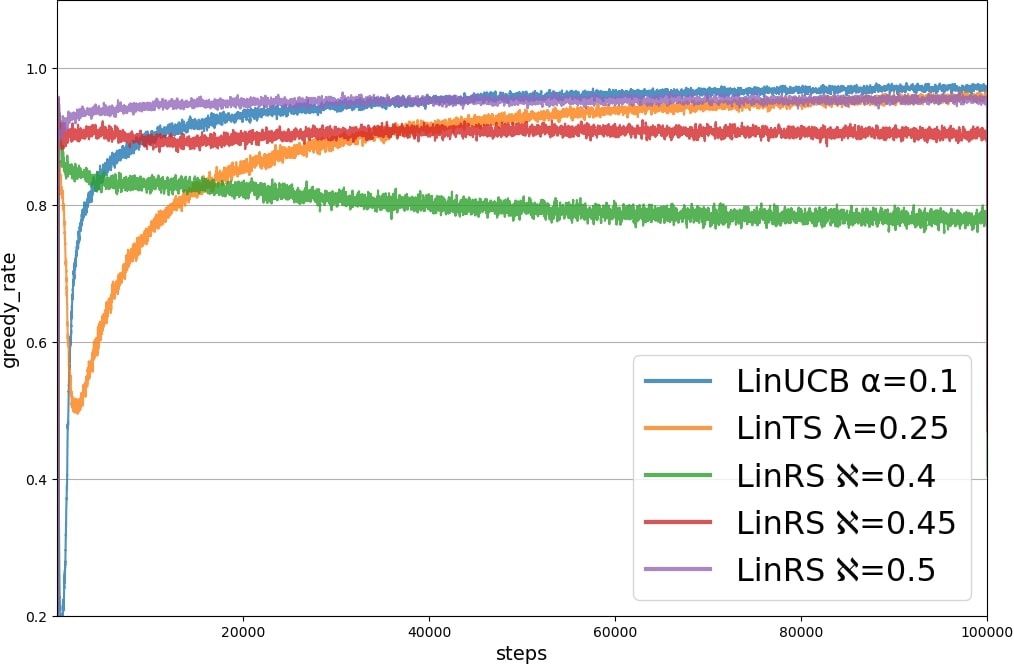}
    	\subcaption{greedy rate: $\aleph_\mathrm{opt}$ = 0.5}
    	\label{greedy_0.5}
  	\end{minipage} \\
  	\\
  	\begin{minipage}[t]{0.49\hsize}
    	\centering
    	\includegraphics[keepaspectratio, scale=0.2]{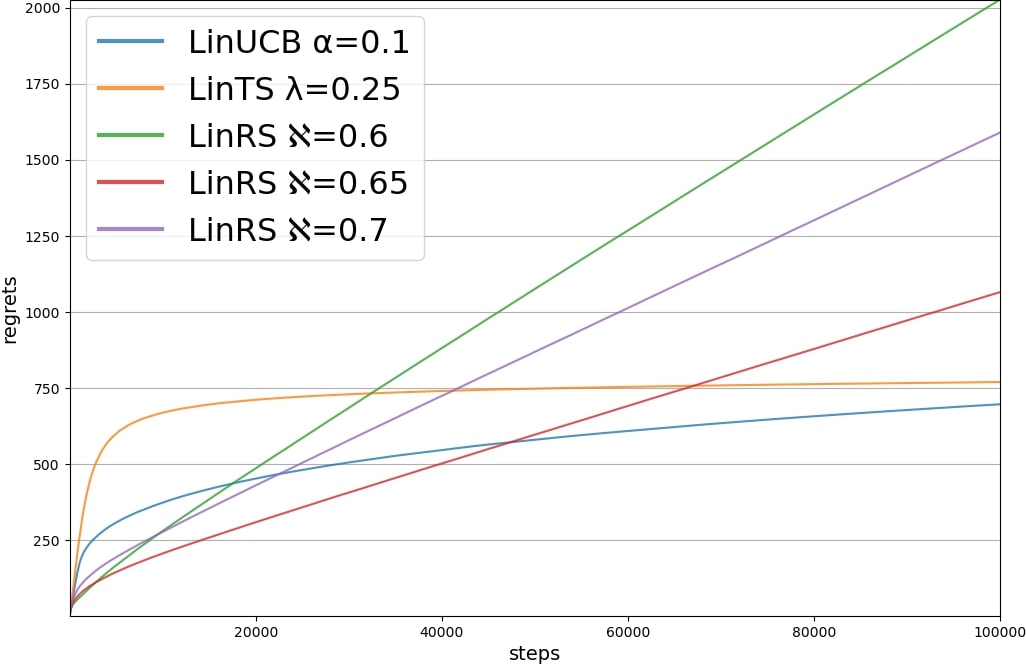}
    	\subcaption{regret: $\aleph_\mathrm{opt}$ = 0.7}
    	\label{regret_0.7}
  	\end{minipage} &
  	\begin{minipage}[t]{0.49\hsize}
    	\centering
    	\includegraphics[keepaspectratio, scale=0.2]{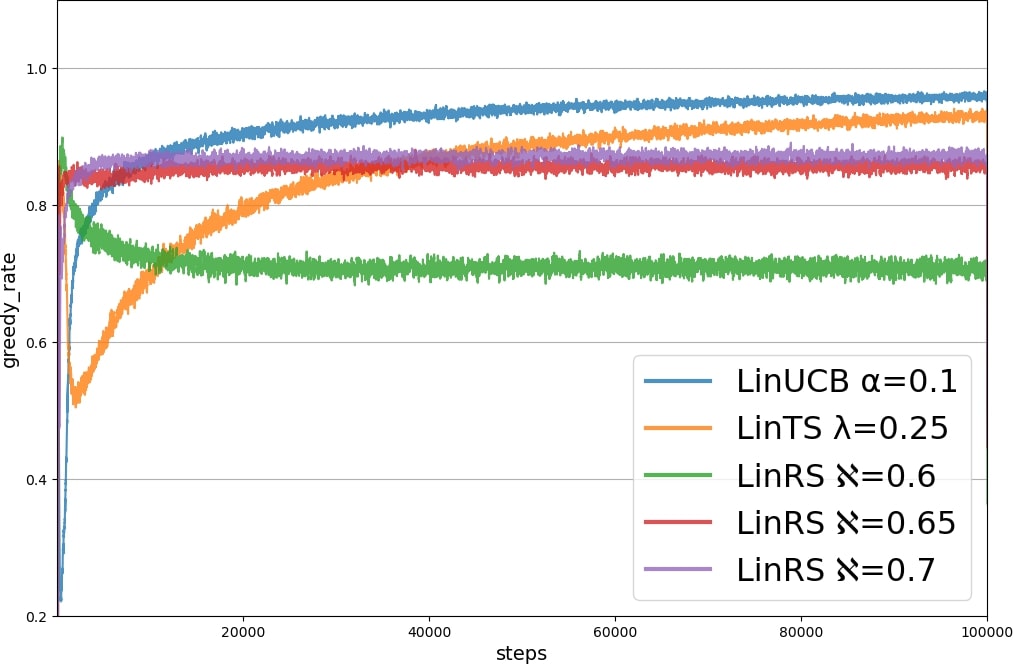}
    	\subcaption{greedy rate: $\aleph_\mathrm{opt}$ = 0.7}
    	\label{greedy_0.7}
  	\end{minipage} \\
\\
  	\begin{minipage}[t]{0.49\hsize}
    	\centering
    	\includegraphics[keepaspectratio, scale=0.2]{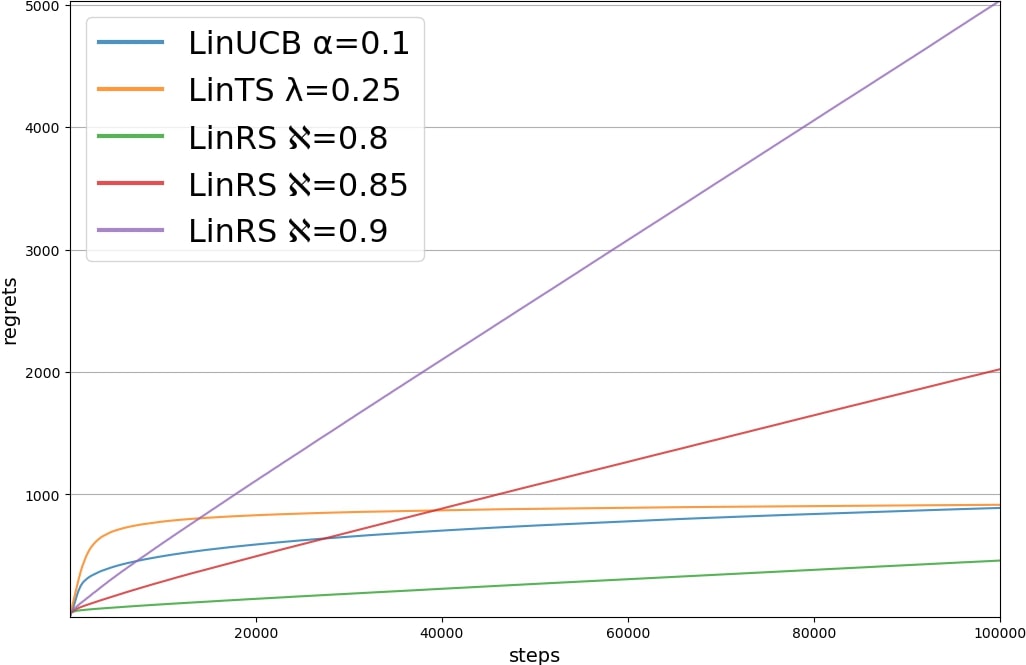}
    	\subcaption{regret: $\aleph_\mathrm{opt}$ = 0.9}
    	\label{regret_0.9}
  	\end{minipage} &
  	\begin{minipage}[t]{0.49\hsize}
    	\centering
    	\includegraphics[keepaspectratio, scale=0.2]{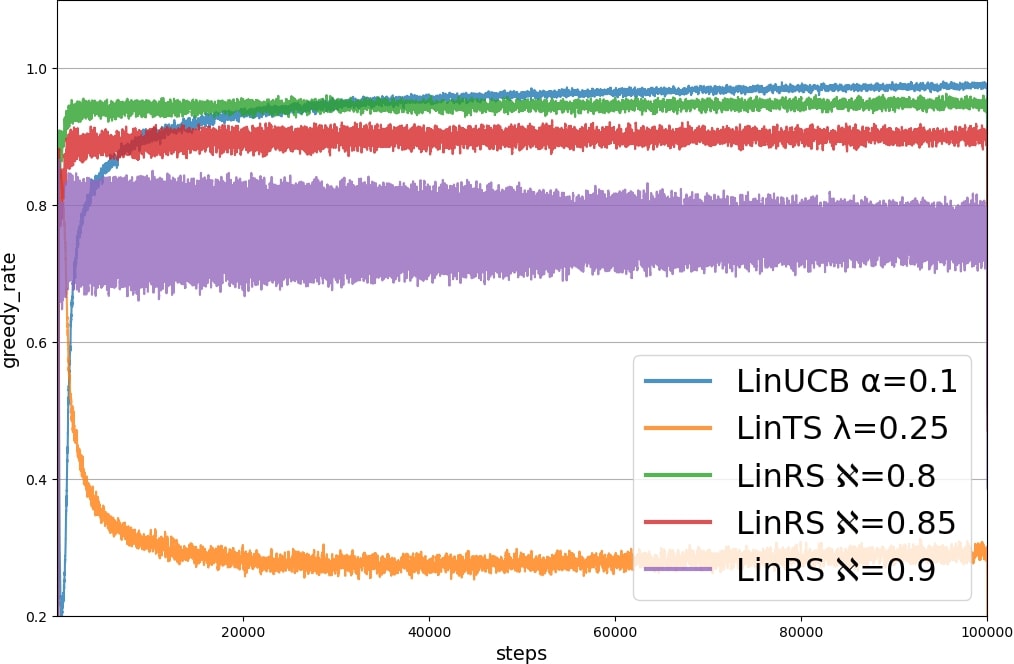}
    	\subcaption{greedy rate: $\aleph_\mathrm{opt}$ = 0.9}
    	\label{greedy_0.9}
  	\end{minipage}
	\end{tabular}
 	\caption{The regret and greedy rates of each algorithm in the experiments with the three artificial datasets ($\aleph_\mathrm{opt} = \{0.5, 0.7, 0.9\}$).  \subref{regret_0.5}
\subref{regret_0.7}
\subref{regret_0.9} is the result of regret.  \subref{greedy_0.5}
\subref{greedy_0.7}
\subref{greedy_0.9} is the result of greedy rate. }
	\label{fig:artificial}
  \end{figure}

Figure \ref{fig:artificial} shows the regret and greedy rates of each algorithm in the experiments with three artificial datasets.
 
  \noindent For the dataset $\aleph_\mathrm{opt} = \{0.5, 0.9\}$, LinRS shows lower regret than the two existing algorithms.
In addition, LinRS shows lower regret in the early steps than the existing algorithms.
In all three datasets, LinRS shows a high greedy rate from the early steps.
 
Table \ref{table1} shows the average run time per simulation of each algorithm and the ratio of each algorithm compared to the average run time of LinUCB.
\begin{table}
\centering
\caption{Average run time per simulation of each algorithm and ratio of each algorithm compared to the average run time of LinUCB.}
\label{table1}
\begin{tabular}{ccc}
\toprule
\textbf{algorithm} 	& \textbf{average run time (s)} & \textbf{ratio}\\
\midrule
LinUCB		& \multicolumn{1}{r}{40.712} & \multicolumn{1}{r}{1.000}\\
LinTS		& \multicolumn{1}{r}{176.855} & \multicolumn{1}{r}{4.344}\\
LinRS		& \multicolumn{1}{r}{49.892} & \multicolumn{1}{r}{1.225}\\
\bottomrule
\end{tabular}
\end{table}
\noindent LinRS is approximately 1.225 times more time consuming than LinUCB to calculate the decision and is approximately 4.344 times more time consuming than LinUCB.
 
\section{Experiments with Real-World Datasets}
\label{sec:real_data}
 
In this section, we evaluate the performance of the proposed LinRS algorithm on the contextual bandit problem using two real-world datasets.
Real-world datasets are more difficult to estimate than artificial datasets because linear approximation is not guaranteed.
By evaluating the performance of the proposed algorithm on a real-world dataset, we verified whether it can be applied to real-world tasks.
\subsection{Real-world datasets}
\label{subsec:real_data_sets}
 
We used the Mushroom and Jester datasets as real-world datasets.
The reward was set based on the study of Riquelme et al. \cite{LinTS}.
\subsubsection{Mushroom dataset}
 
The Mushroom dataset \cite{mushroom_data} consists of two classes: edible mushrooms and poisonous mushrooms.
This dataset has 22 features, such as umbrellas, handles, and habitats.
By representing these features using the One-Hot vectors, the dimensions of the feature vector become $d = 117$. 
The total number of data is $n = 8124$, including 4208 edible mushrooms and 3916 poisonous mushrooms, and the number of actions is $k = 2$. 
Table \ref{mushroom_reward} shows the reward settings for the Mushroom dataset.

\begin{table}
\centering
\caption{Reward settings for the Mushroom dataset. Five rewards are awarded for eating edible mushrooms, five positive rewards with a probability of 0.5, and otherwise a large negative reward of $-35$  for eating poisonous mushrooms.
When the agent does not eat the mushroom, the reward is zero, regardless of the type of mushroom.}
\label{mushroom_reward}
\begin{tabular}{ccc}
\toprule
 & \multicolumn{2}{c}{action}\\
\cmidrule(r){2-3}
	& eat & no eat\\
\midrule
edible mushrooms		& \multicolumn{1}{c}{+5} & \multicolumn{1}{c}{0}\\
poisonous mushrooms	& \multicolumn{1}{c}{+5 or -35 (probability of 0.5)} & \multicolumn{1}{c}{0}\\
\bottomrule
\end{tabular}
\end{table}
\noindent When the agent eats the poisonous mushroom, the agent is randomly given either a positive reward or a large negative reward. 
The reason for giving a positive reward is that there is a possibility that eating the poisonous mushroom will not cause any physical disorder. 
However, as the expected reward for eating the poisonous mushroom is $-15$, not eating the poisonous mushroom is the correct action. 

By using the Mushroom dataset, we created a contextual bandit problem wherein an agent decides whether to eat a given mushroom.
Specifically, the agent judges whether a given mushroom is edible or poisonous based on the given features. 
For example, if the agent mistakenly judges an edible mushroom to be poisonous based on the features, the agent chooses not to eat it and thus receives no reward, and the value is $0$. 
Conversely, if the agent mistakenly judges a poisonous mushroom to be edible, the agent chooses to eat it, and the expected reward is $-15$. 
\subsubsection{Jester dataset}
 
The Jester dataset \cite{jester_data} is a dataset containing each user's assessment of 40 different jokes. 
The total data are n = 19181. 
Among the 40 jokes assessed, 32 were used as features, and the remaining 8 were used as actions.
That is, $d = 32, k = 8$.
The agent recommends one of the eight jokes and obtains the user's assessment of the selected joke as a reward based on 32 jokes.
The assessment of the joke was given as a continuous value from $-10$ to 10.
The feature and reward for each action are thus also continuous values from $-10$ to 10.

Using the Jester dataset, we created a contextual bandit problem for a recommender system in which the agent recommends jokes that are likely to be preferred by users. 
For example, there is a user who appreciates a classic joke among the 32 jokes. 
If the agent recommends a classic joke out of 8 jokes to this user, the evaluation value (reward) will be close to 10. 
Conversely, if the agent recommends an ironic joke, a high evaluation is not expected.
 
\subsection{Experimental setup}
\label{subsec:real_data_setup}
 
We conducted experiments using the Mushroom and Jester datasets.
We ran 8000 steps for the Mushroom dataset, and 10000 steps for the Jester dataset.
The average regret and greedy rate over 100 simulation repetitions are the results.
As mentioned in Section \ref{sec:artificial_data}, regret is used as a measure the balance between exploration and exploitation of the algorithm, and the greedy rate is used as an intuitive measure to check the autonomy of the exploration. 
The agent selected all actions 10 times immediately after the start of the simulation to update the parameters of each algorithm.
We set $\mathrm{batch size} = 20$ for all the algorithms.
Because the parameter update of the reliability of LinRS does not use a closed form, we set $\mathrm{epoch} = 5$ only for the parameter update of the reliability.
The maximum length of the queue of LinRS is set to 100.
\subsection{Algorithm}
\label{subsec:real_data_algo}
 
The proposed algorithm, LinRS, was verified and compared with the existing algorithms, namely, LinUCB and LinTS, which are commonly used in contextual bandit problems.
We tuned the parameters of all the algorithms.
The proposed algorithm, LinRS, was compared and verified with the existing algorithms, LinUCB, and LinTS, which are used in contextual bandit problems.
We tuned the parameters of all the algorithms.
 	 \begin{itemize}
  	\item \textbf{LinUCB} : This algorithm is an extension of UCB to linear functions.
It selects an action using a value function, which is the unbiased estimate of the action value plus $\alpha$ times the unbiased estimate of the variance.
In this experiment, we set the parameter $\alpha = 0.1$.
\item \textbf{LinTS} : This algorithm is an extension of TS to linear functions.
The reward expectation parameter $\hat{\boldsymbol \theta}_a$ is generated from a multivariate normal distribution using vectors and matrices calculated from $\lambda$.
The variance scaling parameter $\sigma^2$ is generated from the inverse gamma distribution using $a_t, b_t$.
In this experiment, we set the parameters as $\lambda = 0.25, a_0 = b_0 = 6$.
\item \textbf{LinRS} : This is the proposed algorithm shown in Algorithm \ref{alg1}.
In the experiment using the Mushroom dataset, we set the aspiration level $\aleph = 4.0$, whereas we set the aspiration level to $\aleph = 2.0$ in the experiment using the Jester dataset.
The parameters required for learning the reliability were $w = \eta = 0.1$ for the Mushroom dataset and $w = \eta = 0.01$ for the Jester dataset.
  	\end{itemize}
\subsection{Results}
\label{subsec:real_data_result}
\begin{figure}[htbp]
	\begin{tabular}{cc}
  	\begin{minipage}[t]{0.49\hsize}
    	\centering
    	\includegraphics[keepaspectratio, scale=0.2]{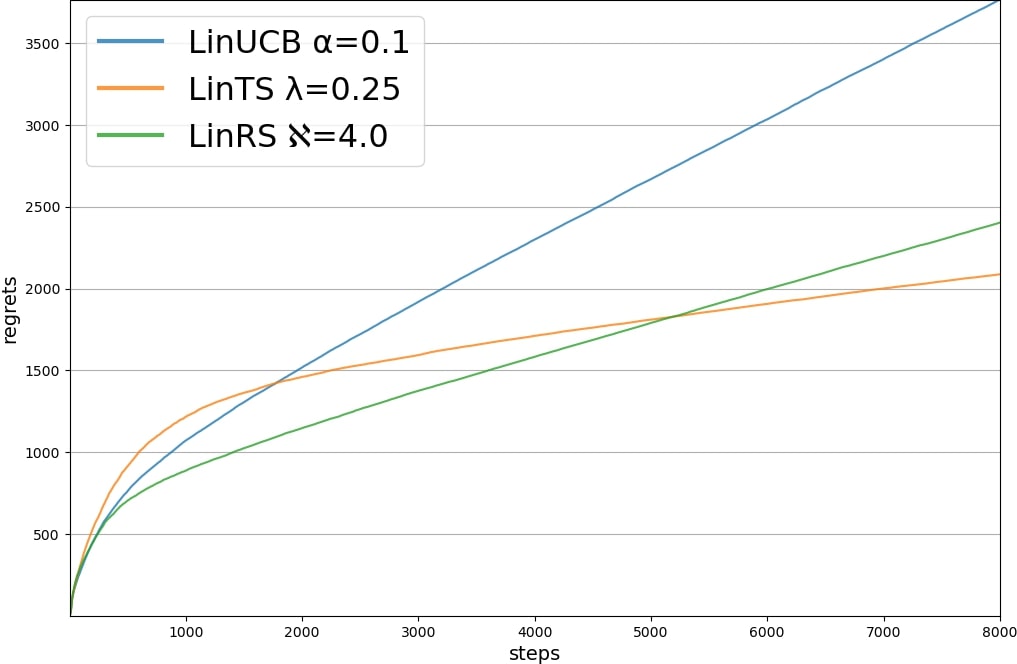}
    	\subcaption{regret: Mushroom}
    	\label{regret_mushroom}
  	\end{minipage} &
  	\begin{minipage}[t]{0.49\hsize}
    	\centering
	\includegraphics[keepaspectratio, scale=0.2]{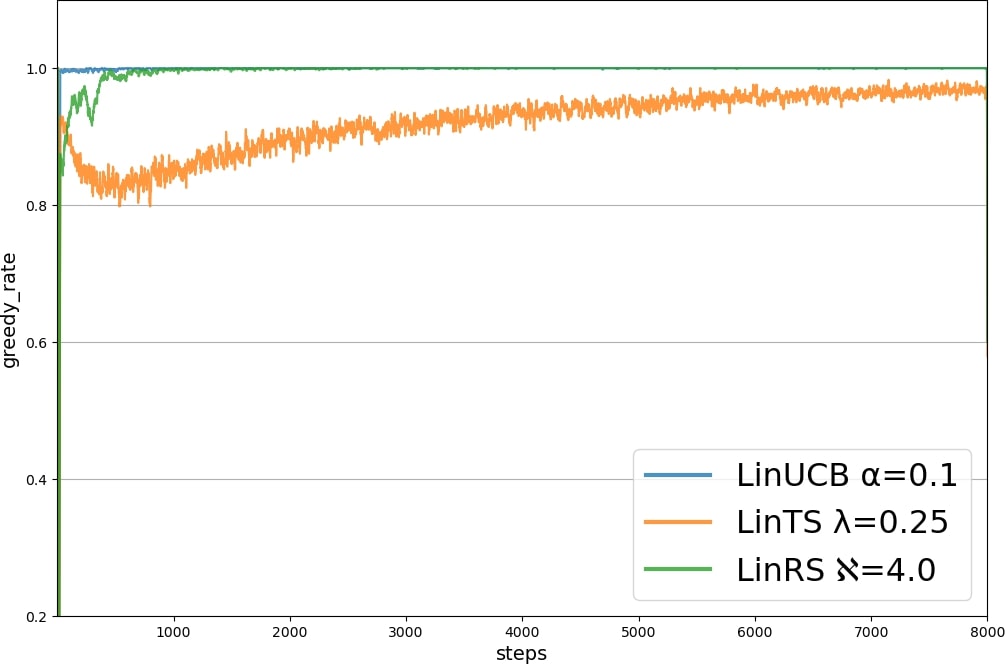}
    	\subcaption{greedy rate: Mushroom}
    	\label{greedy_mushroom}
  	\end{minipage} \\
\\
  	\begin{minipage}[t]{0.49\hsize}
    	\centering
    	\includegraphics[keepaspectratio, scale=0.2]{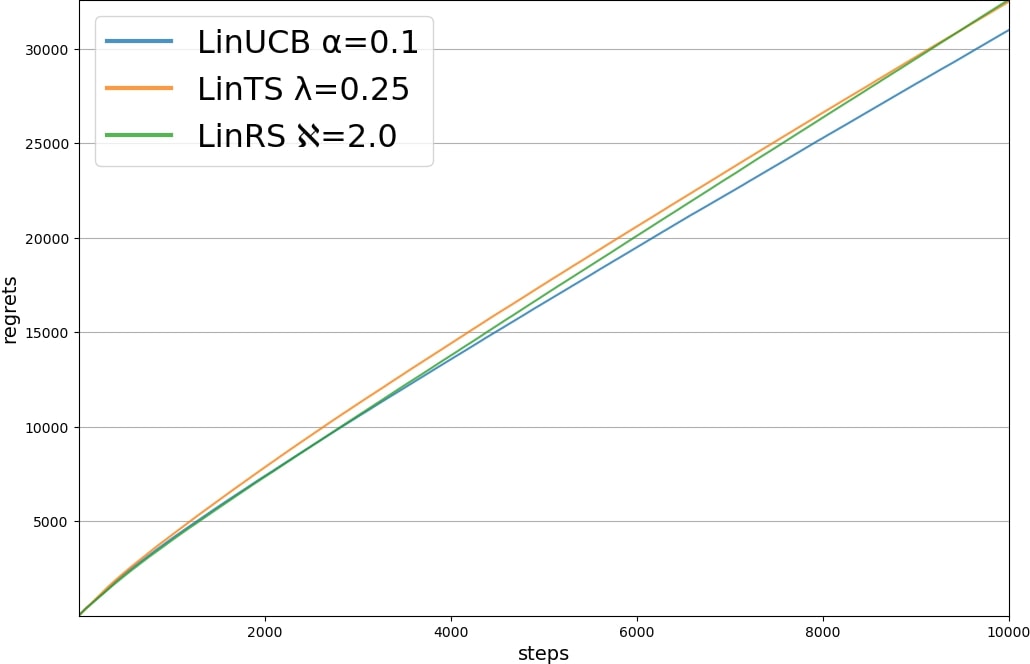}
    	\subcaption{regret: Jester}
    	\label{regret_jester}
  	\end{minipage} &
  	\begin{minipage}[t]{0.49\hsize}
    	\centering
    	\includegraphics[keepaspectratio, scale=0.2]{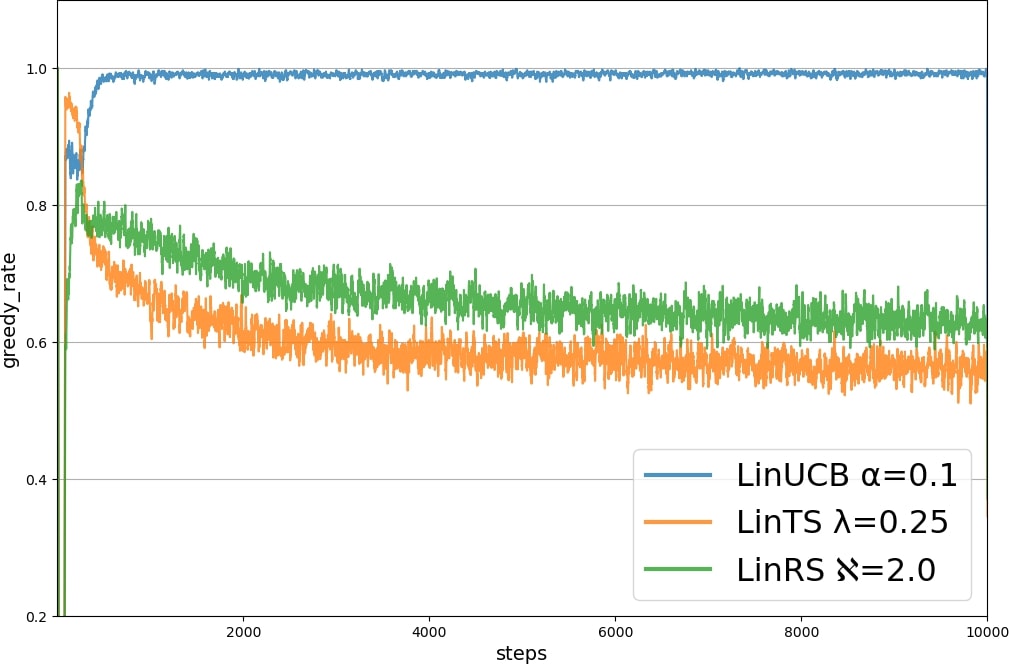}
    	\subcaption{greedy rate: Jester}
    	\label{greedy_jester}
  	\end{minipage}
	\end{tabular}
 	\caption{Regret and greedy rates of each algorithm in experiments with two different real-world datasets (i.e., Mushroom and Jester). \subref{regret_mushroom} \subref{regret_jester} is the result of regret. \subref{greedy_mushroom}
\subref{greedy_jester} is the result of greedy rate. }
	\label{fig:real}
  \end{figure}

Figure \ref{fig:real} shows the regret and greedy rates of each algorithm in experiments with the two different real-world datasets.
 
\noindent In the experiment using the Mushroom dataset, LinRS showed a lower regret than LinUCB.
In addition, LinRS showed lower regret than LinTS in the early steps.
As in the results of the artificial dataset, LinRS showed a high greedy rate from the early steps.
In the experiment using the Jester dataset, the regret of all algorithms linearly increased and showed high regret.

\section{Discussion}
\label{sec:dis}
 
In this section, we discuss our hypotheses and the properties of the LinRS.
\subsection{Efficient learning by reducing the number of explorations and calculation time}
\label{subsec:dis1}
 
LinRS can learn satisfactory actions efficiently while reducing the number of explorations.
In all results except for Figure \ref{regret_jester}, LinRS shows a low regret from the early steps.
In particular, in Figure \ref{regret_0.9}, LinRS always showed lower regret than the existing optimization algorithms.
In addition, in the results of Figure \ref{greedy_0.5}, \ref{greedy_0.7}, \ref{greedy_0.9} and \ref{greedy_mushroom}, LinRS shows a high greedy rate from the early steps.
This indicates that LinRS reduces the number of explorations by switching to exploitation when satisficing actions are being discovered.
In contrast, the results in Figure \ref{regret_0.7}, \ref{regret_0.9}, \ref{greedy_0.7} and \ref{greedy_0.9} show that when a high aspiration level is given, which is not immediately satisfactory, there is a problem that the agent cannot begin exploiting sufficiently early.
In Figure \ref{greedy_0.7} and \ref{greedy_0.9}, LinRS with the optimal aspiration level showed a slower increase in the greedy rate than LinRS with the aspiration level set lower than the optimal aspiration level.
The higher optimal aspiration levels in Figure \ref{regret_0.7},\ref{greedy_0.7}, \ref{regret_0.9} and \ref{greedy_0.9} compared to those in Figure \ref{regret_0.5} and \ref{greedy_0.5} suggest that the discovery of satisfactory action may have been delayed.
 
From the results in Figure \ref{greedy_0.5}, \ref{greedy_0.7}, \ref{greedy_0.9}, \ref{greedy_mushroom}, and \ref{greedy_jester}, we can see that LinUCB exploits, similarly to LinRS, the early steps.
LinUCB shows excellent results even when the data are sparse \cite{LinUCB}; that means it was able to demonstrate less regret while reducing the number of explorations even in the early steps when the data are still small.
Although LinUCB and LinRS have similar characteristics in the early steps, LinRS is relatively superior because it can reduce more the number of explorations and regrets.
 
In addition, the results in Table \ref{table1} show that the calculating time of LinRS is approximately 1.225 times longer than that of LinUCB.
The two algorithms have many similarities, such as the linear approximation of the action value and the calculation of the value function.
This probably can explain why the run times of the two algorithms are so close.
One difference between the two algorithms, though, is the approximation of reliability.
LinUCB uses an unbiased estimator of variance, which plays a similar role to the reliability, and it is defined by the same variables as the unbiased estimator of action value.
Meanwhile, the reliability of LinRS requires two new variables to be defined.
Because LinRS requires more variables than LinUCB and takes more time for calculations, we can assume that the overall calculation is slower.
LinTS samples the parameters of the action value from the posterior distribution and causes it to take approximately four times as long as LinRS and LinUCB.
In real-world tasks where the contextual bandit problem is applied, we can assume that many of the tasks require quick decision-making.
In such tasks, LinRS and LinUCB are superior to LinTS.
\subsection{Advantages of real-world tasks}
\label{subsec:dis3}
 
The results suggest that LinRS is superior to existing optimization algorithms, particularly for short-term tasks.
In the results of Figures \ref{regret_mushroom} and \ref{greedy_mushroom}, LinRS shows low regret from the early steps while reducing the number of explorations.
In the results shown in Figure \ref{regret_jester}, however, all the algorithms failed to show practical performance.
The reason for the difference in the results of the two real-world datasets is the presence or lack of class labels.
In the Mushroom data, the reward distribution is determined by the class.
The fact that the reward distributions of mushrooms in the same class are strongly correlated with each other makes it relatively easy for agents to estimate the action value.
However, it is difficult to classify Jester data clearly in the same manner with Mushroom data.
Furthermore, as the reward distribution of Jester data is not determined depending on the class, the correlation of the reward distribution among users is weak.
These are, we think, the reasons why it is difficult for the agents to utilize the reward information obtained in the past and it is difficult to estimate the action value.
 
In the experiments using Jester data, all algorithms failed to show good performance.
Initially, this result might seem to conclude that the application of linear approximation methods to recommendation systems is difficult.
However, this is not necessarily true.
In the Jester dataset, we used the assessments of 32 jokes as the user's feature vector.
However, actual recommendation systems in the real world use personal data, including the user's age and occupation, as features in addition to the assessment.
Such types of user's personal information has a significant impact on the reward distribution. When combined with the feature vector would, it can improve the accuracy of the action value estimation.
In addition, algorithms can be applied to the task of maximizing the click-through rate by giving a reward of 0 or 1, rather than the assessment from $-10$ to 10.
By adjusting the design of features and rewards, linear approximation methods including LinRS can be fully applied to real-world tasks.
 
\subsection{Comparison of the RS and LinRS properties}
LinRS retains the property of RS that it can learn satisfactory actions efficiently while reducing the number of explorations.
LinRS can thus be said to be a successful linear approximation extension of the RS.
In contrast, LinRS shows the tendency that its optimal aspiration level differs from that of RS owing to the influence of the approximation error.
The results in Figures \ref{regret_0.7} and \ref{regret_0.9} show that the optimal aspiration level of LinRS does not always follow equation (\ref{aleph_opt}).
Linear approximation methods are more prone to make errors in the estimation of the action value than the method without approximation, because of the effect of approximation error.
These errors are a major problem in the decision-making of LinRS, which decides whether an action is satisfied based on the result of the difference between the action value and aspiration level.
To address this problem, in the future study, it is necessary to either reduce the approximation error of the action value or to reduce the effect of the error by making the judgment of satisfactory or-not of an action probabilistic.
Because of the unavoidableness of error, the satisficing criterion set as per the equation (\ref{aleph_opt}) may not always be optimal, it is possible to provide the optimal aspiration level by proper tuning.
Figure \ref{regret_0.5}, \ref{regret_0.9}, \ref{regret_mushroom} and \ref{regret_jester} shows that LinRS with the optimal aspiration level shows the same or lower regret than the existing methods.
This implies that by tuning the aspiration level, LinRS can perform as well as or better than the existing methods.
 
\section{Conclusions}
\label{sec:con}
 
In this paper, we proposed a satisficing algorithm LinRS with linear models.
We assessed the performance of LinRS on contextual bandit problems using artificial and real-world datasets.
The results show that LinRS can efficiently learn satisfactory actions with a small amount of exploration while reducing the run time for action selection.
This suggests that satisficing algorithms may be suitable for use in more complex environments, such as those used in deep reinforcement learning.
 
Satisficing algorithms are likely to be more advantageous than optimizing algorithms in complex environments where the problem of the large number of explorations is more pronounced.
Generalization of LinRS leads to a function approximation method for an RS-based satisficing algorithm, applicable to deep reinforcement learning by utilizing the findings in this study.
The results of the experiments on artificial datasets showed that LinRS does not always set the optimal aspiration level by using the same equation as RS.
This may be caused by the influence of an approximation error.
In this study, the aspiration level ($\aleph$) was set in advance. A method for autonomously finding the optimal aspiration level for LinRS based on a theoretical analysis makes LinRS more practical in applications (for the most basic case, see \cite{RS_aleph}).
Devising a method to set the optimal aspiration level of the generalized LinRS leads to a solution of the same type of problem that is expected to happen in function approximation methods.
Extensions of RS are to be expected to accelerate the application of reinforcement learning algorithms in real-world tasks with complex environments.

\section*{Acknowlegments}

This work was partially supported by JSPS KAKENHI Grant Number 20H04259. 

\section*{Author contribution}
Conceptualization, A.M., Y.K. and T.T.; methodology, A.M. and Y.K; software, A.M.; validation, A.M.; formal analysis, A.M.; investigation, A.M.; resources, T.T.; data curation, A.M.; writing---original draft preparation, A.M.; writing---review and editing, Y.K. and T.T.; visualization, A.M.; supervision, Y.K. and T.T.; project administration, Y.K.; funding acquisition, T.T. All authors have read and agreed to the published version of the manuscript.

\section*{Data availability}
Publicly available datasets were analyzed in this study. The datasets and source files are found here: [\url{https://github.com/akane-minami/linrs}].

\section*{conflictsofinterest}
The authors declare no conflict of interest.

\bibliographystyle{model1-num-names}
\bibliography{sample.bib}





\end{document}